\pdfoutput=1

\documentclass{article}

\usepackage[preprint]{wenxiaobai}

\usepackage[utf8]{inputenc}
\usepackage[T1]{fontenc}
\usepackage{hyperref}
\usepackage{url}
\usepackage{booktabs}
\usepackage{amsfonts}
\usepackage{amsmath}
\usepackage{amssymb}
\usepackage{microtype}
\usepackage{xcolor}
\usepackage{atbegshi}

\usepackage{float}
\usepackage{placeins}
\usepackage{capt-of}
\usepackage{caption}
\usepackage{tabularx}
\usepackage{graphicx}
\usepackage{multirow}
\usepackage{subfigure}
\usepackage{overpic}
\usepackage{enumitem}
\usepackage{wrapfig}
\graphicspath{{figures/}}

\newcommand{\bench}{\textit{WildGraphBench}}

\title{WildGraphBench: Benchmarking GraphRAG with Wild-Source Corpora}

\author{
  \bf Pengyu Wang$^{1*}$, Benfeng Xu$^{1,2,\dagger}$, Licheng Zhang$^{1,\S}$, Shaohan Wang$^{1}$,\\\bf Mingxuan Du$^{1}$, Chiwei Zhu$^{1}$, Zhendong Mao$^{1}$ \\[0.5em]
  $^{1}$University of Science and Technology of China, Hefei, China \\
  $^{2}$Metastone Technology, Beijing, China \\[0.3em]
  \texttt{\{wangpengyu, benfeng, zlczlc\}@mail.ustc.edu.cn, zdmao@ustc.edu.cn}
}

\makeatletter
\renewcommand{\@noticestring}{\rule{0.4\linewidth}{0.4pt}\\[0.5em]$^*$Work done during the internship at Metastone.\\$^\dagger$Project lead.\\$^\S$Corresponding author.\\Preprint. Work in progress.}
\makeatother

\begin{document}

\maketitle

\begin{abstract}
\hspace{2em}Graph-based Retrieval-Augmented Generation (GraphRAG) organizes external knowledge as a hierarchical graph, enabling efficient retrieval and aggregation of scattered evidence across multiple documents.
However, many existing benchmarks for GraphRAG rely on short, curated passages as external knowledge, failing to adequately evaluate systems in realistic settings involving long contexts and large-scale heterogeneous documents.
To bridge this gap, we introduce \bench, a benchmark designed to assess GraphRAG performance in the wild. 
We leverage Wikipedia's unique structure, where cohesive narratives are grounded in long and heterogeneous external reference documents, to construct a benchmark reflecting real-word scenarios.
Specifically, we sample articles across 12 top-level topics, using their external references as the retrieval corpus and citation-linked statements as ground truth, resulting in 1,100 questions spanning three levels of complexity: single-fact QA, multi-fact QA, and section-level summarization. 
Experiments across multiple baselines reveal that current GraphRAG pipelines help on multi-fact aggregation when evidence comes from a moderate number of sources, but this aggregation paradigm may overemphasize high-level statements at the expense of fine-grained details, leading to weaker performance on summarization tasks.
Project page: \url{https://github.com/BstWPY/WildGraphBench}.
\end{abstract}

\section{Introduction}

Retrieval-augmented generation (RAG) grounds LLM outputs by retrieving evidence from an external corpus\citep{lewis2021retrievalaugmentedgenerationknowledgeintensivenlp}, but it may struggle when scattered evidence must be extracted from multiple documents and integrated into a coherent answer.
Recently, Graph-based RAG (GraphRAG) \citep{peng2024graphretrievalaugmentedgenerationsurvey} has gained increasing attention as a paradigm that builds a graph over documents or chunks and performs graph-guided expansion and aggregation for multi-document evidence assembly and long-context reasoning \citep{zhang2025surveygraphretrievalaugmentedgeneration}.

\begin{figure}[!htbp]
  \centering
  \includegraphics[width=0.9\linewidth]{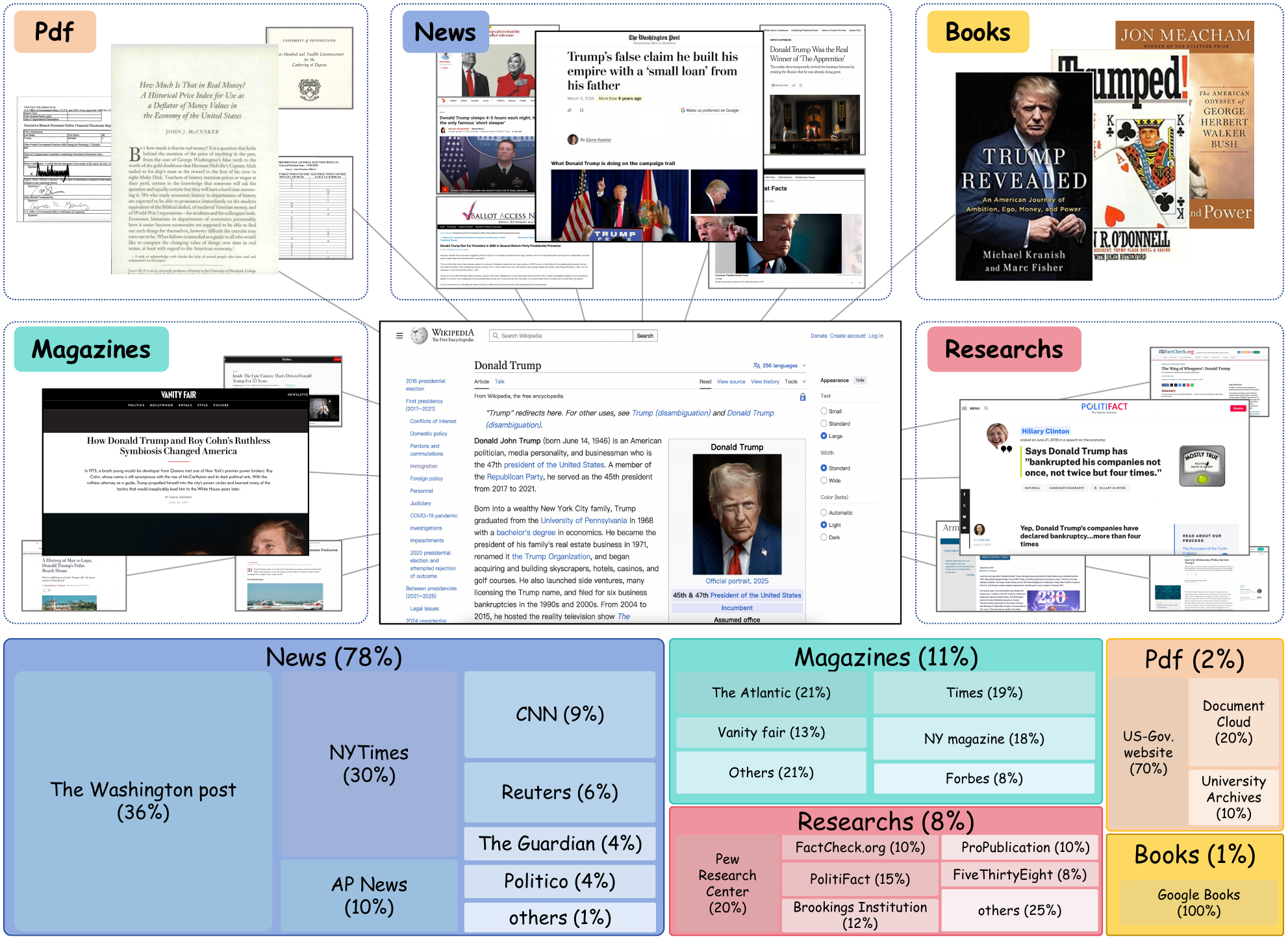}
  \caption{Why we use Wikipedia references as wild evidence. Wikipedia articles are concise summaries with citation-linked statements. The linked reference pages are often long, noisy, and heterogeneous (e.g., news sites, blogs, PDFs, and public reports). This mismatch makes evidence retrieval and verification harder.}
  \label{fig:wild-need}
\end{figure}

Many GraphRAG methods have been proposed, exploring different graph constructions and retrieval strategies.
Microsoft GraphRAG\citep{edge2025localglobalgraphrag} builds document-level graphs and supports local-to-global aggregation for query-focused summarization, while LightRAG\citep{guo2025lightragsimplefastretrievalaugmented} improves evidence coverage by coupling an entity--relation graph with vector retrieval and multi-stage expansion. 
In addition, some works attempt to improve practicality.
Fast-GraphRAG\citep{fastgraphrag2024} focuses on efficiency and adopts a lightweight graph retrieval pipeline to reduce indexing and querying costs.
HippoRAG2\citep{gutiérrez2025ragmemorynonparametriccontinual} further extends evidence access by introducing an external knowledge graph and filtering mechanism.
LinearRAG\citep{zhuang2025linearraglineargraphretrieval} enhances the practicality by adopting a lightweight hierarchy and propagation-style ranking for scalable indexing and retrieval.
These strategies enable structured evidence expansion and aggregation, allowing LLMs to address more complex queries.

However, current GraphRAG benchmarks still rely on short, curated passages.
As a result, retrieval and generation in long-context settings with large-scale, heterogeneous document collections remain under-tested.
Yet this setting is central to real-world applications and is also where GraphRAG is expected to be most beneficial.
In addition, many benchmarks make the task closer to lookup-and-stitch: once a few pre-trimmed passages are retrieved, a system can answer by simple concatenation or light paraphrasing rather than true multi-document aggregation.
Traditional datasets such as HotpotQA \citep{yang2018hotpotqadatasetdiverseexplainable}, 2WikiMultiHopQA \citep{xanh2020_2wikimultihop}, and MultiHop-RAG \citep{tang2024multihopragbenchmarkingretrievalaugmentedgeneration} typically follow this assumption.
Even some recent benchmarks extend to longer domain documents, e.g., UltraDomain \citep{qian2025memoragboostinglongcontext}, they still assume cleaner document boundaries and less heterogeneous sources than wild corpora.
For GraphRAG specifically, a few dedicated benchmarks have recently been proposed \citep{xiao2025graphragbenchchallengingdomainspecificreasoning,xiang2025usegraphsragcomprehensive}, providing controlled protocols and corpora; however, their corpora remain more structured and less heterogeneous than in-the-wild settings.
This gap calls for tasks that test reliable aggregation under long contexts and heterogeneous, uncurated sources.

\begin{figure}[t]
  \centering
  \includegraphics[width=0.95\linewidth]{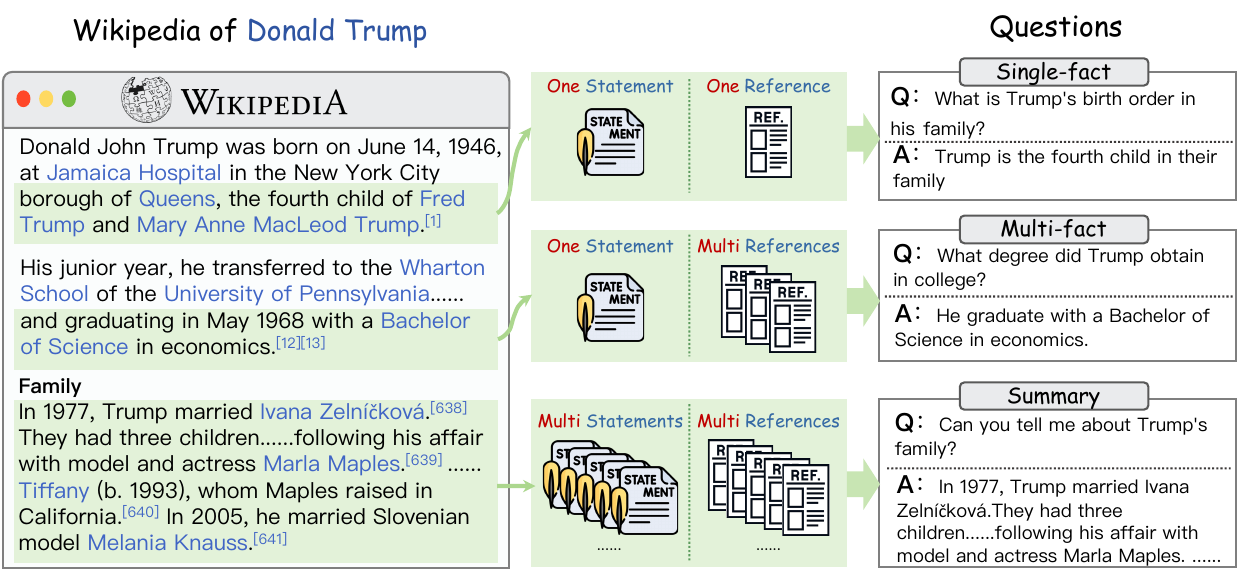}
  \caption{Example instances in \bench: (1) single-fact questions grounded by a single gold statement and one reference, (2) multi-fact questions requiring evidence aggregation across multiple statements/references, and (3) section-level summary questions evaluated at the statement level.}
  \label{fig:task-cases}
\end{figure}

To bridge this gap, we introduce \bench, a benchmark that targets the in-the-wild scenario of GraphRAG, where a system is expected not only to retrieve evidence from long, heterogeneous corpora, but also to synthesize an answer whose correctness depends on assembling scattered support across multiple sources rather than a handful of pre-trimmed passages.
We instantiate this setting using Wikipedia: each article provides a concise entry with citation-linked statements, while its external reference pages form a long, heterogeneous web corpus (Figure~\ref{fig:wild-need}).
We therefore sample Wikipedia articles from 12 top-level topics, using each article's reference pages as the retrieval corpus and treating citation-linked Wikipedia statements as ground-truth facts.
We create 1,100+ questions in three types, as shown in Figure~\ref{fig:task-cases}, spanning single-fact lookup, multi-fact evidence aggregation, and section-level summarization, which together stress the spectrum from precise retrieval to broad factual coverage.
Our contributions are as follows:
\begin{itemize}
    \item We construct \bench, a dataset based on Wikipedia references that reflects real-world noise and complexity.
        \item We design three question types: single-fact, multi-fact, and section-level summary.Then we introduce a statement-grounded evaluation method for single-fact, multi-fact, and summary questions.
    \item Experiments on multiple methods show that GraphRAG improves multi-fact aggregation but struggles with broad summary tasks.
\end{itemize}

\section{Related Work}

\paragraph{Retrieval-Augmented Generation.}
Retrieval-augmented generation (RAG) answers a query by retrieving related text from an external corpus, then generating based on that text \citep{lewis2021retrievalaugmentedgenerationknowledgeintensivenlp}. A common setup uses a retriever (e.g., BM25) plus an LLM reader \citep{INR-019,chen-etal-2017-reading}. Prior work shows retrieval can reduce hallucination compared to direct generation, especially when evidence is needed \citep{shuster2021retrievalaugmentationreduceshallucination}.

\paragraph{Graph Retrieval-Augmented Generation.}
Graph retrieval-augmented generation (GraphRAG) builds a graph over chunks or documents and retrieves evidence through graph operations, aiming to capture complex dependencies that flat retrieval might miss.
Microsoft GraphRAG \citep{edge2025localglobalgraphrag} introduces a modular pipeline that uses community detection algorithms to generate hierarchical summaries, supporting both local entity queries and global thematic answering.
LightRAG \citep{guo2025lightragsimplefastretrievalaugmented} constructes a dual-level entity--relation graph and couples it with vector retrieval, allowing for multi-stage evidence expansion across low-level entities and high-level concepts.
Focusing on efficiency, Fast-GraphRAG \citep{fastgraphrag2024} adopts a lightweight design with optimized indexing strategies to significantly reduce the computational overhead of graph maintenance.
HippoRAG2 \citep{gutiérrez2025ragmemorynonparametriccontinual} draws inspiration from the human brain, introducing an external knowledge graph and a memory-style retrieval mechanism powered by Personalized PageRank (PPR) to filter noise and discover multi-hop paths.
Furthermore, LinearRAG \citep{zhuang2025linearraglineargraphretrieval} targets large-scale scalability by utilizing a linear-complexity propagation ranking method, overcoming the bottleneck of traditional graph algorithms.
These methods collectively demonstrate that structured graph traversal can significantly enhance evidence aggregation compared to flat baselines.

\paragraph{Benchmarks for RAG and GraphRAG.}
Several benchmarks study retrieval and multi-hop reasoning under different evidence settings.
HotpotQA\citep{yang2018hotpotqadatasetdiverseexplainable}, 2WikiMultiHopQA\citep{xanh2020_2wikimultihop}, and MultiHop-RAG\citep{tang2024multihopragbenchmarkingretrievalaugmentedgeneration} focus on multi-hop question answering.
UltraDomain\citep{qian2025memoragboostinglongcontext} evaluates retrieval and generation over longer domain corpora.
GraphRAG-Bench\citep{xiao2025graphragbenchchallengingdomainspecificreasoning} provides corpora and protocols tailored to graph-based retrieval.
Additionally, Xiang et al. \citep{xiang2025usegraphsragcomprehensive} systematically analyze the scenarios where graph structures provide a clear benefit over flat retrieval.
Despite these advances, there remains a gap for a benchmark that simultaneously stresses long-context processing, noise robustness, and multi-document aggregation in a wild, unstructured web setting.

\section{WildGraphBench}
\label{sec:benchmark}

\begin{figure}[t]
  \centering
  \includegraphics[width=0.95\linewidth]{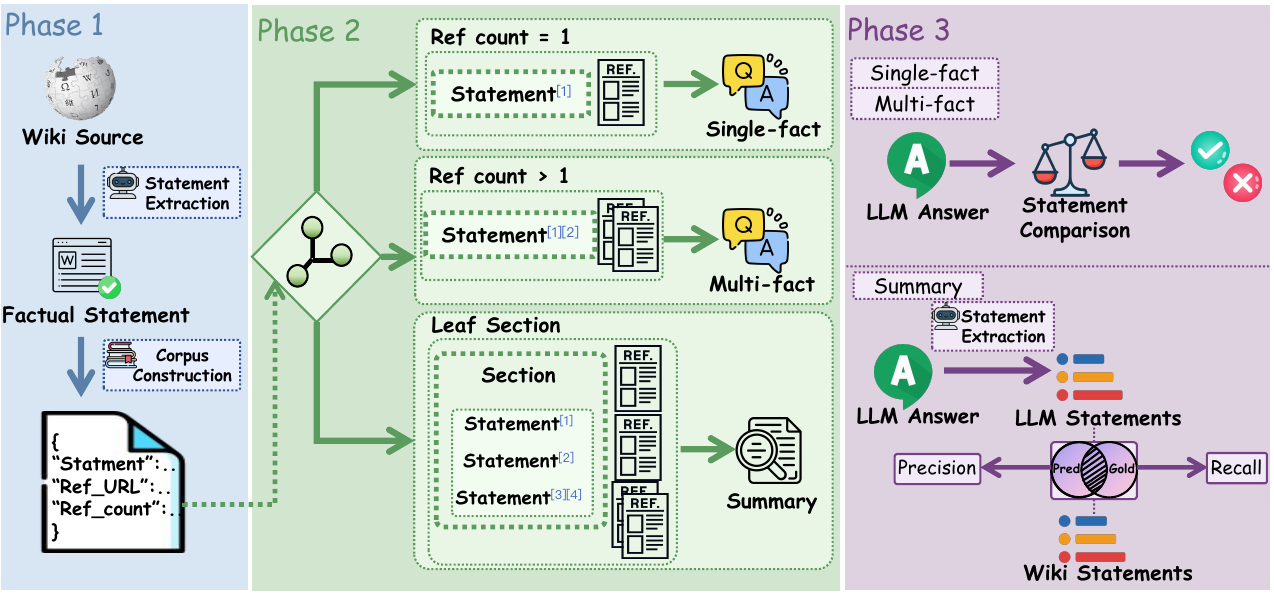}
  \caption{Three-phase workflow of \bench\ after data collection. Phase~1:citation-aware statement extraction, producing the Wikipedia gold corpus. Phase~2: design single-fact, multi-fact, and section-level summary questions. Phase~3: evaluate with statement-grounded accuracy and statement-level precision/recall/F1.}
  \label{fig:pipeline}
\end{figure}

In this section, we describe how \bench\ is constructed.
As shown in Figure~\ref{fig:pipeline}, the framework has three phases.
First, we collect reference pages and extract citation-linked statements from Wikipedia leaf sections, producing the Wikipedia gold corpus.
Based on the extracted corpus, we then build three types of questions: single-fact, multi-fact, and section-level summary.
Finally, we introduce a statement-grounded evaluation method.

\subsection{Data Collection}
\label{sec:data-collection}

We start from 12 high-level Wikipedia topics\footnote{
Source: Wikipedia:Contents.
It lists 13 top-level topics: Culture, Geography, Health, History, Human activities, Mathematics, Nature, People, Philosophy, Religion, Society, Technology, and Reference.
We drop \textit{Reference}.
It mainly describes how to write and cite Wikipedia pages, rather than a content domain.
} and within each topic select articles with a large number of references, as these articles tend to have dense and diverse citation structures. For each article, we collect all reference URLs and fetch the original web pages using jina.ai.\footnote{\url{https://jina.ai}}. Crucially, if the original page fails but an archive exists, we use the archived page to ensure data completeness. We keep the raw page text, including boilerplate and noise, to simulate the wild retrieval environment.

\subsection{Statement Extraction}
\label{sec:seg-extract}

We use citation-linked Wikipedia statements as the ground-truth factual units for evaluation.
.To extract statements, we first split each Wikipedia article into leaf sections using a simple regex parser over Wiki markup. Each leaf section has a section path (e.g., \textit{Donald Trump > Political Career > Impeachments}). In each leaf section, we identify sentences containing citation markers. For each sentence, an LLM rewrites it into a clean factual statement by removing footnote markers and fixing local coreference issues. We also parse the Wiki markup to retrieve the exact reference URLs.

\subsection{Wikipedia Gold Corpus Construction}
\label{sec:corpus-construction}

We align the references by matching Wikipedia reference URLs to the crawled pages. If a cited sentence is missing any referenced page, we drop it to ensure quality. The resulting Wikipedia gold corpus is organized at the granularity of leaf sections, where each section stores a list of triples:
\begin{equation}
\mathcal{T} = (\mathit{statement}, \mathit{ref\_urls}, \mathit{ref\_count})
\label{eq:triple}
\end{equation}
where $statement$ is the normalized factual statement, $ref\_urls$ is the set of reference URLs associated with the original cited sentence, and $ref\_count$ is the number of references. This corpus serves as the authoritative source of gold statements for evaluation.

\begin{table}[!htbp]
    \centering
    \caption{Statistics of Question Types in \bench. The dataset consists of 1,197 questions distributed across three distinct categories.}
    \label{tab:question_stats}
    {\normalsize
    \setlength{\tabcolsep}{20pt}
    \begin{tabular}{c c}
        \toprule
        \textbf{Question Type} & \textbf{Count} \\
        \midrule
        Single-Fact & 667 \\
        Multi-Fact  & 191 \\
        Summary     & 339 \\
        \midrule
        \textbf{Total} & \textbf{1,197} \\
        \bottomrule
    \end{tabular}
    }
\end{table}

\subsection{Question Design}
\label{sec:question-design}

We design three types of questions on top of the Wikipedia gold corpus. The reference count $ref\_count$ attached to each Wikipedia triple determines whether it is used for a single-fact or multi-fact question, while leaf sections themselves serve as the basis for summary questions (Table~\ref{tab:question_stats}).

\paragraph{Single-fact questions.}
If a triple has $ref\_count=1$, we use it to generate a single-fact question. Given the article title, the section path, the original source sentence, and the cleaned statement, we prompt an LLM to write a question whose answer is exactly that statement (up to minor paraphrasing). The prompt specifically encourages the model to include multiple constraints (e.g., entity, time, and location) and discourages copying long spans from the statement, ensuring that the question is non-trivial but still tightly aligned with the gold statement and its supporting evidence.

\paragraph{Multi-fact questions.}
If a triple has $ref\_count\geq 2$, we treat it as a candidate for a multi-fact question. Intuitively, these statements are supported by multiple references and often describe relationships that span several sources. We again condition an LLM on the article title, section path, and statement to generate a question that requires recovering that statement. In addition, we enforce a strict multi-reference check: for each such triple, we ask an LLM judge whether any single reference alone is sufficient to support all key facts in the statement, and only keep those triples for which at least two references are jointly required. This yields questions that genuinely test a model's ability to aggregate evidence from multiple documents.

\paragraph{Section-level summary questions.}
For summary questions, we operate at the level of leaf sections. For a given leaf section, we collect all
valid triples under that section and deduplicate their statements to obtain the gold statement set $S^\ast$. 
We then prompt an LLM to generate a natural information-seeking question based on the article title and the section path, but not on the section text itself, so that the question is phrased independently of the exact wording of Wikipedia. The expected answer to such a question is the set $S^\ast$, i.e., the factual content of the leaf section. During evaluation, systems are required to retrieve and summarize information under noisy, long-context evidence conditions.

\subsection{Evaluation Metrics}
\label{sec:metrics}

We evaluate systems differently for single-fact/multi-fact questions and for summary questions, but in all cases the gold answer is defined at the level of factual statements.

\paragraph{Single-fact and multi-fact accuracy.}
For single-fact and multi-fact questions, each instance is associated with exactly one gold statement $s^\ast$. Given a system answer $\hat{a}$ and the relevant evidence, an LLM judge decides whether $\hat{a}$ is factually equivalent to $s^\ast$ under the evidence. We assign a score of 1 if the answer is correct and 0 otherwise, and report accuracy separately for single-fact and multi-fact questions.

\paragraph{Statement-level score for summary.}
For summary questions, the gold answer is the statement set $S^\ast$ extracted from the corresponding leaf section.
Given a system output, we run a statement extractor to obtain a set of predicted statements $\hat{S}=\{\hat{s}_1,\dots,\hat{s}_k\}$.
We then define a binary matching function $\mathit{Match}(s,\hat{s})\in\{0,1\}$, which returns 1 if $\hat{s}$ is a correct paraphrase of $s$ (i.e., it conveys the same fact), and 0 otherwise.
Using this match, we compute statement-level recall and precision as:
\begin{gather}
\mathrm{Recall} = \frac{1}{|S^\ast|}\sum_{s\in S^\ast}\max_{\hat{s}\in \hat{S}}\mathit{Match}(s,\hat{s})
\label{eq:recall}\\
\mathrm{Precision} = \frac{1}{|\hat{S}|}\sum_{\hat{s}\in \hat{S}}\max_{s\in S^\ast}\mathit{Match}(s,\hat{s})
\label{eq:precision}
\end{gather}
The F1 score is the harmonic mean of precision and recall.
This metric directly measures factual coverage (recall) and hallucination rate (precision), while remaining robust to minor paraphrasing.

\section{Experiments}
\label{sec:experiments}

\subsection{Experimental Settings}
\label{sec:exp-setting}

\paragraph{Implementation Details}
For methods requiring pre-chunking, we segment documents with chunk size 1200 tokens and overlap 100 tokens.
At retrieval time, we set $\texttt{top\_k}=5$ for single-fact and multi-fact questions, and $\texttt{top\_k}=10$ for summary questions.
To align generation and graph construction across systems, we use \texttt{gpt-4o-mini} as the default model for graph construction and answering.
For evaluation, we use \texttt{gpt-5-mini} as the LLM judge to score single-fact and multi-fact accuracy and to compute statement-level precision/recall/F1 for summary questions.

\paragraph{Evaluated Methods}
We evaluate \bench\ on representative flat-RAG and GraphRAG-style baselines.
For flat-RAG, we include NaiveRAG \citep{lewis2021retrievalaugmentedgenerationknowledgeintensivenlp} and BM25 \citep{INR-019}.
For GraphRAG-style methods, we evaluate Fast-GraphRAG \citep{fastgraphrag2024}, Microsoft GraphRAG (local/global) \citep{edge2025localglobalgraphrag}, LightRAG (hybrid) \citep{guo2025lightragsimplefastretrievalaugmented}, LinearRAG \citep{zhuang2025linearraglineargraphretrieval}, and HippoRAG2 \citep{gutiérrez2025ragmemorynonparametriccontinual}.

\subsection{Main Results}

\begin{table}[!htbp]
\centering
\caption{Main results on \bench. While graph-based methods show clear advantages on multi-fact questions requiring aggregation, flat baselines like NaiveRAG remain competitive on single-fact retrieval and achieve higher recall on summary tasks due to broader context coverage.}
\label{tab:main-results}
\small
\renewcommand{\arraystretch}{1.15}
\begin{tabular*}{\linewidth}{@{\extracolsep{\fill}}ccccccc}
\toprule
\multirow[c]{2}{*}{\raisebox{-0.5ex}{\textbf{Method}}} &
\multicolumn{3}{c}{\textbf{Question Answering}} &
\multicolumn{3}{c}{\textbf{Summary}} \\
\cmidrule(lr){2-4}\cmidrule(lr){5-7}
& \textbf{Avg. Acc.} & \textbf{Single-fact Acc.} & \textbf{Multi-fact Acc.} & \textbf{Recall} & \textbf{Precision} & \textbf{F1} \\
\midrule
NaiveRAG & 59.79 & 66.87 & 35.08 & \textbf{13.54} & 19.07 & \textbf{15.84} \\
BM25 & 36.83 & 41.38 & 20.94 & 9.38 & 19.46 & 12.66 \\
\midrule
Fast-GraphRAG & 33.56 & 35.83 & 25.65 & 6.81 & 23.48 & 10.56 \\
HippoRAG2 & \textbf{64.33} & \textbf{71.51} & 39.27 & 11.15 & 16.76 & 13.39 \\
Microsoft GraphRAG(local) & 38.23 & 39.43 & 34.03 & 9.82 & 12.64 & 11.05 \\
Microsoft GraphRAG(global) & 54.54 & 56.52 & \textbf{47.64} & 12.66 & 15.13 & 13.78 \\
LightRAG(hybrid) & 56.76 & 61.32 & 40.84 & 12.44 & 17.7 & 14.61 \\
LinearRAG & 44.87 & 47.53 & 35.6 & 5.81 & \textbf{29.2} & 9.69 \\
\bottomrule
\end{tabular*}
\end{table}

Table~\ref{tab:main-results} reports the overall performance on \bench\  and the  \texttt{people} subset.
We observe a consistent pattern across both parts.

On the relatively simple \textbf{single-fact} questions, flat retrieval-augmented baselines remain highly competitive.
In particular, NaiveRAG achieves strong accuracy, outperforming most graph-based variants on the Novel Dataset; among the GraphRAG-style methods, only HippoRAG2 attains a higher single-fact accuracy (71.51 vs.\ 66.87), suggesting that graph retrieval does not automatically translate into gains when the answer can often be supported by a single salient chunk.
BM25 is also competitive, and in some cases surpasses several graph methods, indicating that keyword matching remains a strong prior for straightforward fact lookup.

The advantage of graph-based retrieval becomes more visible on harder tasks.
For \textbf{multi-fact} questions, which require aggregating evidence from multiple references, Microsoft GraphRAG(global) achieves the best accuracy(47.64), and several graph variants are comparable to or better than NaiveRAG and BM25.
This implies that structured traversal / global context aggregation can help when evidence is scattered and must be combined, while a purely flat top-$k$ pipeline is more likely to miss complementary pieces of information.

\begin{table}[!htbp]
\centering
\caption{Results on the \texttt{people} subset of \bench, compared with human performance.}
\label{tab:people-results}
\small
\renewcommand{\arraystretch}{1.15}
\begin{tabular*}{\linewidth}{@{\extracolsep{\fill}}ccccccc}
\toprule
\multirow[c]{2}{*}{\raisebox{-0.5ex}{\textbf{Method}}} &
\multicolumn{3}{c}{\textbf{Question Answering}} &
\multicolumn{3}{c}{\textbf{Summary}} \\
\cmidrule(lr){2-4}\cmidrule(lr){5-7}
& \textbf{Ave. Acc.} & \textbf{Single-fact Acc.} & \textbf{Multi-fact Acc.} & \textbf{Recall} & \textbf{Precision} & \textbf{F1} \\
\midrule
NaiveRAG & 65.82 & 76.62 & 28.12 & \textbf{10.48} & 15.29 & \textbf{8.03} \\
BM25 & 65.2 & 74.03 & 34.38 & 5.74 &16.98 &5.03 \\
\midrule
Fast-GraphRAG & 30.43 & 33.77 & 18.75 & 1.48 & \textbf{22.83} & 1.62 \\
HippoRAG2 & 64.89 & 72.73 & 37.5 & 7.63 & 15.69 & 6.14 \\
Microsoft GraphRAG(local) & 35.16 & 38.96 & 21.88 & 4.59 & 9.17 & 2.98 \\
Microsoft GraphRAG(global) & 56.81 & 62.34 & 37.5 & 5.52& 14.13 & 5.41 \\
LightRAG(hybrid) & \textbf{74.42} & \textbf{80.52} & \textbf{53.12} & 5.56 & 15.69 & 4.73 \\
LinearRAG & 45.26 & 51.95 & 21.88 & 1.52 & 22.51 & 1.69\\
\midrule
Human performance & 85.66 & 89.61 & 71.88 & 38.59 & 12.62 & 15.3 \\
\bottomrule
\end{tabular*}
\end{table}

For \textbf{summary} questions, all methods obtain low statement-level scores, highlighting the difficulty of reconstructing a leaf section's factual content from long, noisy evidence.
Notably,  \textbf{NaiveRAG achieves the highest recall and the best F1} on our Dataset (Table~\ref{tab:main-results}).
A plausible explanation is that summary questions demand \emph{broad coverage}: re
trieving a wider variety of raw evidence chunks can directly increase the chance that the generator sees more gold facts, improving recall and thus F1.
In contrast, many GraphRAG-style systems introduce additional bottlenecks---entity/relation extraction, graph sparsification, neighborhood summarization, and traversal budgets---which may degrade under web noise and long contexts.
When graph construction is imperfect (missing entities/edges) or when traversal/summarization budgets are limited, these methods can fail to \emph{scale} their evidence gathering to the breadth required by section-level summaries, resulting in lower recall and weaker overall F1 even if they sometimes improve precision via filtering.
This suggests that scaling GraphRAG to summary-style tasks under wild evidence requires more robust graph construction and higher-capacity aggregation, rather than relying on graph structure alone.

Overall, these results indicate that \textbf{GraphRAG is not always advantageous on easy questions}---it can be more expensive than NaiveRAG or BM25 without clear gains for single-fact lookup.
While GraphRAG shows promising improvements on multi-fact aggregation, summary questions remain challenging under noisy long-context evidence, and method design must carefully balance \emph{coverage} (recall) versus \emph{filtering} (precision) under a fixed retrieval/compute budget.

\subsection{Graph Analysis}

\begin{figure}[!htbp]
  \centering
  \includegraphics[width=0.7\linewidth]{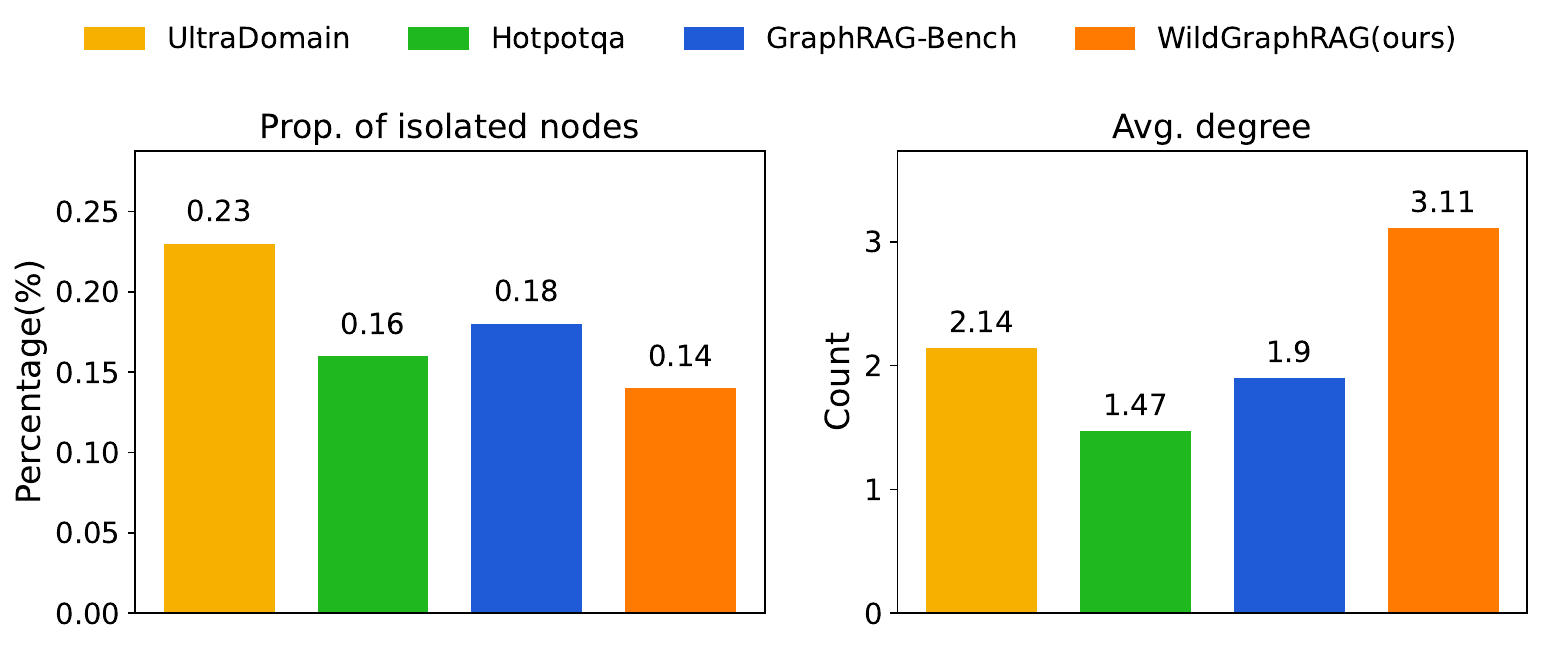}
  \caption{\textbf{Graph Quality.} We build graphs with LightRAG. Left: the fraction of isolated nodes (lower means better connectivity). Right: the average degree (higher means denser links).}
  \label{fig:graph-quality}
\end{figure}

Figure~\ref{fig:graph-quality} and Table~\ref{tab:graph-stats} compare graph connectivity across datasets under the same LightRAG construction pipeline.
Following \citet{xiang2025usegraphsragcomprehensive}, a more \emph{organized} graph should avoid excessive isolated nodes, because graph-based retrieval (e.g., traversal, community aggregation, or PPR-style propagation) relies on connectivity to reach and combine evidence beyond a single chunk.
Consistent with this criterion, our graph achieves the \textbf{highest average degree} (3.11) and the \textbf{lowest proportion of isolated nodes} (0.14), indicating denser cross-document links and fewer disconnected components.

In contrast, the other datasets are structurally less favorable for graph connectivity under the same pipeline.

\begin{table}[!htbp]
\centering
\caption{Graph structural statistics across datasets constructed using the LightRAG pipeline. \bench\ exhibits a significantly higher max degree and denser connectivity, reflecting the complex, hub-centric nature of wild reference pages.}
\label{tab:graph-stats}
\small
\renewcommand{\arraystretch}{1.1}
\begin{tabular*}{\linewidth}{@{\extracolsep{\fill}}ccccc}
\toprule
\textbf{Dataset} &
\textbf{Tokens} &
\textbf{Nodes} &
\textbf{Degrees} &
\textbf{Max Degree}\\
\midrule
UltraDomain &15.1M&33018&35284&195 \\
HotpotQA&1.4M&28282&20762&50 \\
GraphRAG-Bench&1.1M&11150&10638&155\\
Ours &2.4M&23940&\textbf{37246}&\textbf{967} \\
\bottomrule
\end{tabular*}
\end{table}

\textbf{UltraDomain} is constructed from curated long-context domain corpora (e.g., textbooks and domain documents), where content is organized by chapters/sections with cleaner boundaries and less repeated entity co-occurrence across documents; as a result, entity mentions are more ``local'', producing more low-degree or isolated nodes.

\textbf{HotpotQA} provides short, paragraph-level Wikipedia evidence for multi-hop QA; such contexts are comparatively compact and often concentrate on answering a specific question with a small set of supporting paragraphs, leading to weaker global entity sharing and thus a sparser graph.

\textbf{GraphRAG-Bench} exhibits structural sparsity due to domain-specific characteristics across its subsets. The \textbf{Novel} subset features narrative-style text where entities and events are often distributed with strong locality (e.g., confined to specific scenes or chapters), and coreference-heavy writing dilutes explicit entity overlap during extraction. Similarly, the \textbf{Medical} subset—composed largely of biomedical guidelines from high terminological density and distinct, non-narrative structures. The lack of explicit continuity between independent medical documents isolates entities within their specific contexts, while the extraction ambiguity of specialized terms (e.g., drug names) by general-purpose LLMs further exacerbates graph fragmentation, resulting in lower average degrees and more disconnected components across the benchmark.

Finally, Table~\ref{tab:graph-stats} highlights a striking \textbf{max-degree} gap: although the UltraDomain is much larger in total token budget than ours, our graph still exhibits a dramatically larger max degree.
This suggests the presence of \emph{hub entities} that are repeatedly linked by many distinct pages, i.e., multiple sources converge on the same entity-centric node.
Such hub-and-spoke patterns make the benchmark substantially harder for retrieval and generation: systems must aggregate partially overlapping evidence from many documents and synthesize a coherent answer, directly stressing cross-document multi-source summarization and reasoning---the key capability that \bench\ aims to evaluate.

\subsection{Human Performance}
\label{sec:human-performance}

We recruit domain-knowledgeable annotators (graduate-level or above) and ask them to answer questions under the same evidence constraint as RAG systems.
Interestingly, we observe a distinct behavior in human responses for summary tasks: human annotators tend to prioritize comprehensive coverage of key facts, attempting to include as many details as possible.
While this approach sometimes results in lower precision—as models are often more conservative in their generation—the overall F1 score for human performance remains high (e.g., 15.30 F1 on people subset, see Table~\ref{tab:people-results}), effectively serving as a strong upper-bound reference for the difficulty of evidence aggregation in our benchmark.

\subsection{Ablation Study}
\label{sec:sensitivity}
\begin{wrapfigure}{r}{0.4\linewidth}
  \centering
  \vspace{-0.5cm}
  \includegraphics[width=\linewidth]{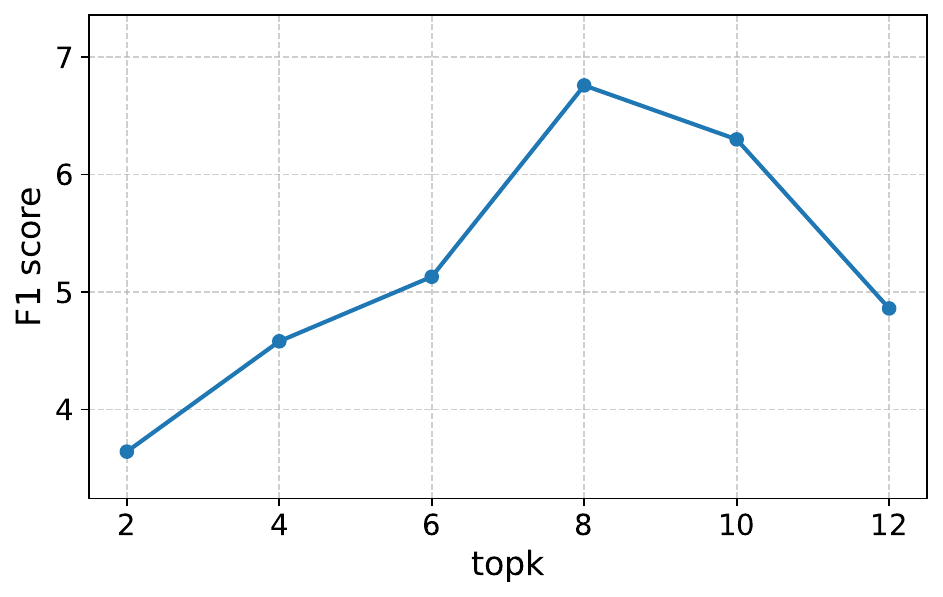}
  \caption{Impact of retrieval budget (top-$k$) on F1 score for summary questions. F1 increases as $k$ grows, then drops when $k$ is too large. It peaks at $k=8$.}
  \label{fig:topk-analysis}
\end{wrapfigure}
We further investigate the impact of retrieval budget (top-$k$) on the performance of summary questions. We conduct an experiment using HippoRAG2 on a specific domain subset of our dataset, varying the top-$k$ parameter from 2 to 12. As illustrated in Figure~\ref{fig:topk-analysis}, the F1 score exhibits an inverted U-shaped trend: it initially increases as $k$ grows, reaching a peak at $k=8$, and subsequently declines as $k$ increases further to 12.

This trend suggests that an optimal retrieval budget is crucial for balancing recall and precision. When top-$k$ is too small (e.g., $k < 5$), the system fails to retrieve sufficient evidence chunks to cover the broad factual content required for summarization, limiting recall. Conversely, when top-$k$ becomes too large (e.g., $k > 8$), the introduction of excessive irrelevant noise or "distractor" chunks overwhelms the generator's context window or reasoning capability, leading to hallucinations or loss of focus, which degrades the overall F1. This finding emphasizes that retrieval strategies must be carefully tuned to the corpus size and noise level to maximize performance on complex summary tasks.

\section{Conclusions}

We introduce \bench, a benchmark designed to evaluate GraphRAG on wild-source corpora. Utilizing the heterogeneous reference pages cited in Wikipedia as the retrieval corpus while grounding statements in the Wikipedia articles themselves, we construct three progressively challenging task types to stress-test retrieval, aggregation, and summarization in uncurated environments.
Our experiments reveal that while graph-based retrieval offers limited gains over strong flat baselines for simple, single-fact queries, it demonstrates significant advantages on multi-fact questions requiring cross-document evidence aggregation. Conversely, performance on summarization tasks remains low across all methods in this wild setting, highlighting the critical need for more robust evidence acquisition and synthesis mechanisms in real-world scenarios.

\section*{Limitations}
\label{sec:limitations}

Our benchmark derives gold statements from Wikipedia, which reflects editorial consensus rather than absolute truth; consequently, the gold set may inherit omissions or inaccuracies from Wikipedia and its citations. 
In addition, our evaluation relies on LLM-based judgment and statement matching, which may introduce systematic biases (e.g., preference for certain phrasing or verbosity) and may not perfectly mirror unbiased human assessment. 
These factors should be considered when interpreting absolute scores and fine-grained comparisons between methods.

\section*{Ethical considerations}
\label{sec:ethical}

\paragraph{Data and intended use.}
\bench\ is constructed from Wikipedia articles and their external reference pages; we use citation-linked Wikipedia statements as gold facts and the cited pages as a noisy retrieval corpus.
We respect the original sources' licenses/terms and do not claim ownership over third-party content; any redistribution or derivatives should comply with the original access conditions.
We specify \bench's intended use as \textbf{research-only benchmarking} for retrieval robustness and multi-document evidence aggregation, and we discourage non-research uses unless explicitly permitted by the source conditions.

\paragraph{Risks.}
Because the corpus contains long, heterogeneous web pages, it may include noise, bias, outdated claims, toxic language, or inadvertently exposed sensitive information.
Accordingly, we recommend standard safety practices (e.g., toxicity/PII filtering when appropriate) and emphasize that \bench\ evaluates robustness rather than establishing absolute truth.

\begin{ack}
We thank the anonymous reviewers and our colleagues for their helpful feedback and discussions.
\end{ack}

\newpage

\bibliography{custom}
\bibliographystyle{acl_natbib}

\newpage
\appendix

\section{Dataset Statistics by Domain}
\label{sec:appendix_domain_stats}


\begin{table}[H]
\centering
\caption{Per-domain question counts in \bench.}
\label{tab:domain_question_counts}
\renewcommand{\arraystretch}{1.1}
\begin{tabular*}{\linewidth}{@{\extracolsep{\fill}}ccccc}
\toprule
\textbf{Domain} & \textbf{Single-Fact} & \textbf{Multi-Fact} & \textbf{Summary} & \textbf{Total} \\
\midrule
Culture            & 86 & 37 & 32 & 155 \\
Geography          & 41 & 24 & 33 & 98  \\
Health             & 76 & 19 & 55 & 150 \\
History            & 25 & 1  & 10 & 36  \\
Human activities   & 83 & 13 & 44 & 140 \\
Mathematics        & 21 & 1  & 11 & 33  \\
Nature             & 18 & 0  & 10 & 28  \\
People             & 77 & 32 & 45 & 154 \\
Philosophy         & 46 & 6  & 18 & 70  \\
Religion           & 72 & 4  & 30 & 106 \\
Society            & 66 & 21 & 27 & 114 \\
Technology         & 56 & 33 & 24 & 113 \\
\midrule
\textbf{Total} & \textbf{667} & \textbf{191} & \textbf{339} & \textbf{1,197} \\
\bottomrule
\end{tabular*}
\end{table}

\onecolumn
\section{Results on every domain}
\label{sec:appendix_results}

\renewcommand{\arraystretch}{1.15}
\setlength{\tabcolsep}{0pt}

\begin{table}[H]
\centering
\small
\label{tab:culture}
\caption{Culture}

\begin{tabular*}{\linewidth}{@{\extracolsep{\fill}}ccccccc}
\toprule
\multirow[c]{2}{*}{\raisebox{-0.5ex}{\textbf{Method}}} &
\multicolumn{3}{c}{\textbf{Question Answering}} &
\multicolumn{3}{c}{\textbf{Summary}} \\
\cmidrule(lr){2-4}\cmidrule(lr){5-7}
& \textbf{Ave. Acc.} & \textbf{Single-fact Acc.} & \textbf{Multi-fact Acc.} & \textbf{Recall} & \textbf{Precision} & \textbf{F1} \\
\midrule
NaiveRAG & 56.91 & 67.44 & 32.43 & 12.69 & 24.84 & 9.79 \\
BM25 & 28.46 & 33.72 & 16.22 & 14.58 & 24.95 & 10.62 \\
\midrule
Fast-GraphRAG & 20.32 & 22.09 & 16.22 & 5.73 & 24.64 & 2.48 \\
HippoRAG2 & 51.22 & 62.79 & 24.32 & 11.02 & 19.79 & 6.68 \\
Microsoft GraphRAG (local) & 41.46 & 43.02 & 37.84 & 3.39 & 12.96 & 1.61 \\
Microsoft GraphRAG (global) & 52.04 & 52.33 & 51.35 & 6.77 & 15.54 & 1.80 \\
LightRAG (hybrid) & 46.34 & 52.33 & 32.43 & 14.25 & 21.36 & 6.93 \\
LinearRAG & 29.27 & 32.56 & 21.62 & 4.69 & 25.33 & 4.91 \\
\bottomrule
\end{tabular*}
\end{table}

\begin{table}[H]
\centering
\small
\label{tab:geography}
\caption{Geography}

\begin{tabular*}{\linewidth}{@{\extracolsep{\fill}}ccccccc}
\toprule
\multirow[c]{2}{*}{\raisebox{-0.5ex}{\textbf{Method}}} &
\multicolumn{3}{c}{\textbf{Question Answering}} &
\multicolumn{3}{c}{\textbf{Summary}} \\
\cmidrule(lr){2-4}\cmidrule(lr){5-7}
& \textbf{Ave. Acc.} & \textbf{Single-fact Acc.} & \textbf{Multi-fact Acc.} & \textbf{Recall} & \textbf{Precision} & \textbf{F1} \\
\midrule
NaiveRAG & 60.00 & 70.73 & 41.67 & 10.92 & 22.83 & 7.89 \\
BM25 & 35.38 & 43.90 & 20.83 & 6.84 & 14.62 & 4.81 \\
\midrule
Fast-GraphRAG & 44.62 & 53.66 & 29.17 & 6.00 & 31.52 & 6.29 \\
HippoRAG2 & 73.85 & 85.37 & 54.17 & 8.59 & 18.77 & 6.28 \\
Microsoft GraphRAG (local) & 49.23 & 60.98 & 29.17 & 11.48 & 16.37 & 8.66 \\
Microsoft GraphRAG (global) & 52.31 & 56.10 & 45.83 & 11.36 & 18.68 & 7.54 \\
LightRAG (hybrid) & 58.47 & 60.98 & 54.17 & 15.31 & 22.07 & 9.61 \\
LinearRAG & 60.00 & 65.85 & 50.00 & 5.71 & 38.66 & 5.87 \\
\bottomrule
\end{tabular*}
\end{table}

\begin{table}[H]
\centering
\small
\label{tab:health}
\caption{Health}

\begin{tabular*}{\linewidth}{@{\extracolsep{\fill}}ccccccc}
\toprule
\multirow[c]{2}{*}{\raisebox{-0.5ex}{\textbf{Method}}} &
\multicolumn{3}{c}{\textbf{Question Answering}} &
\multicolumn{3}{c}{\textbf{Summary}} \\
\cmidrule(lr){2-4}\cmidrule(lr){5-7}
& \textbf{Ave. Acc.} & \textbf{Single-fact Acc.} & \textbf{Multi-fact Acc.} & \textbf{Recall} & \textbf{Precision} & \textbf{F1} \\
\midrule
NaiveRAG & 49.47 & 59.21 & 10.53 & 20.34 & 16.01 & 11.15 \\
BM25 & 32.63 & 39.47 & 5.26 & 13.70 & 19.79 & 8.70 \\
\midrule
Fast-GraphRAG & 31.58 & 36.84 & 10.53 & 10.79 & 17.23 & 6.87 \\
HippoRAG2 & 62.10 & 71.05 & 26.32 & 16.59 & 14.23 & 9.12 \\
Microsoft GraphRAG (local) & 31.58 & 34.21 & 21.05 & 15.60 & 10.53 & 7.01 \\
Microsoft GraphRAG (global) & 54.74 & 60.53 & 31.58 & 20.60 & 12.41 & 10.15 \\
LightRAG (hybrid) & 52.63 & 59.21 & 26.32 & 17.73 & 14.23 & 9.90 \\
LinearRAG & 40.00 & 43.42 & 26.32 & 11.88 & 27.70 & 10.95 \\
\bottomrule
\end{tabular*}
\end{table}

\begin{table}[H]
\centering
\small
\label{tab:history}
\caption{History}

\begin{tabular*}{\linewidth}{@{\extracolsep{\fill}}ccccccc}
\toprule
\multirow[c]{2}{*}{\raisebox{-0.5ex}{\textbf{Method}}} &
\multicolumn{3}{c}{\textbf{Question Answering}} &
\multicolumn{3}{c}{\textbf{Summary}} \\
\cmidrule(lr){2-4}\cmidrule(lr){5-7}
& \textbf{Ave. Acc.} & \textbf{Single-fact Acc.} & \textbf{Multi-fact Acc.} & \textbf{Recall} & \textbf{Precision} & \textbf{F1} \\
\midrule
NaiveRAG & 61.54 & 60.00 & 100.00 & 2.50 & 8.88 & 2.58 \\
BM25 & 34.62 & 36.00 & 0.00 & 2.50 & 23.58 & 2.14 \\
\midrule
Fast-GraphRAG & 34.62 & 32.00 & 100.00 & 0.00 & 23.35 & 0.00 \\
HippoRAG2 & 73.08 & 72.00 & 100.00 & 3.33 & 8.06 & 2.07 \\
Microsoft GraphRAG (local) & 46.15 & 44.00 & 100.00 & 0.00 & 17.12 & 0.00 \\
Microsoft GraphRAG (global) & 65.38 & 64.00 & 100.00 & 5.00 & 12.53 & 5.45 \\
LightRAG (hybrid) & 69.23 & 68.00 & 100.00 & 0.00 & 12.49 & 0.00 \\
LinearRAG & 46.15 & 44.00 & 100.00 & 0.00 & 29.25 & 0.00 \\
\bottomrule
\end{tabular*}
\end{table}

\begin{table}[H]
\centering
\small
\label{tab:human_activities}
\caption{Human Activities}

\begin{tabular*}{\linewidth}{@{\extracolsep{\fill}}ccccccc}
\toprule
\multirow[c]{2}{*}{\raisebox{-0.5ex}{\textbf{Method}}} &
\multicolumn{3}{c}{\textbf{Question Answering}} &
\multicolumn{3}{c}{\textbf{Summary}} \\
\cmidrule(lr){2-4}\cmidrule(lr){5-7}
& \textbf{Ave. Acc.} & \textbf{Single-fact Acc.} & \textbf{Multi-fact Acc.} & \textbf{Recall} & \textbf{Precision} & \textbf{F1} \\
\midrule
NaiveRAG & 81.25 & 85.54 & 53.85 & 12.60 & 24.20 & 8.09 \\
BM25 & 50.00 & 51.81 & 38.46 & 10.38 & 20.32 & 6.50 \\
\midrule
Fast-GraphRAG & 36.46 & 37.35 & 30.77 & 6.88 & 20.16 & 5.13 \\
HippoRAG2 & 79.17 & 84.34 & 46.15 & 13.20 & 18.90 & 7.39 \\
Microsoft GraphRAG (local) & 29.17 & 30.12 & 23.08 & 6.99 & 10.74 & 4.41 \\
Microsoft GraphRAG (global) & 59.38 & 61.45 & 46.15 & 15.85 & 19.49 & 9.71 \\
LightRAG (hybrid) & 64.58 & 68.67 & 38.46 & 9.64 & 21.61 & 7.42 \\
LinearRAG & 51.04 & 50.60 & 53.85 & 4.26 & 19.97 & 3.71 \\
\bottomrule
\end{tabular*}
\end{table}

\begin{table}[H]
\centering
\small
\label{tab:mathematics}
\caption{Mathematics}

\begin{tabular*}{\linewidth}{@{\extracolsep{\fill}}ccccccc}
\toprule
\multirow[c]{2}{*}{\raisebox{-0.5ex}{\textbf{Method}}} &
\multicolumn{3}{c}{\textbf{Question Answering}} &
\multicolumn{3}{c}{\textbf{Summary}} \\
\cmidrule(lr){2-4}\cmidrule(lr){5-7}
& \textbf{Ave. Acc.} & \textbf{Single-fact Acc.} & \textbf{Multi-fact Acc.} & \textbf{Recall} & \textbf{Precision} & \textbf{F1} \\
\midrule
NaiveRAG & 45.46 & 42.86 & 100.00 & 31.82 & 27.08 & 20.60 \\
BM25 & 54.54 & 57.14 & 0.00 & 22.73 & 31.36 & 16.06 \\
\midrule
Fast-GraphRAG & 50.00 & 47.62 & 100.00 & 36.36 & 29.13 & 24.97 \\
HippoRAG2 & 54.54 & 57.14 & 0.00 & 27.27 & 32.63 & 20.22 \\
Microsoft GraphRAG (local) & 45.46 & 47.62 & 0.00 & 31.82 & 23.46 & 18.23 \\
Microsoft GraphRAG (global) & 54.54 & 57.14 & 0.00 & 39.39 & 15.48 & 19.20 \\
LightRAG (hybrid) & 59.09 & 61.90 & 0.00 & 40.91 & 20.31 & 22.32 \\
LinearRAG & 59.09 & 61.90 & 0.00 & 31.82 & 33.63 & 22.11 \\
\bottomrule
\end{tabular*}
\end{table}

\begin{table}[H]
\centering
\small
\label{tab:nature}
\caption{Nature}

\begin{tabular*}{\linewidth}{@{\extracolsep{\fill}}ccccccc}
\toprule
\multirow[c]{2}{*}{\raisebox{-0.5ex}{\textbf{Method}}} &
\multicolumn{3}{c}{\textbf{Question Answering}} &
\multicolumn{3}{c}{\textbf{Summary}} \\
\cmidrule(lr){2-4}\cmidrule(lr){5-7}
& \textbf{Ave. Acc.} & \textbf{Single-fact Acc.} & \textbf{Multi-fact Acc.} & \textbf{Recall} & \textbf{Precision} & \textbf{F1} \\
\midrule
NaiveRAG & 50.00 & 50.00 & 0.00 & 13.33 & 23.86 & 6.50 \\
BM25 & 38.89 & 38.89 & 0.00 & 8.33 & 11.87 & 5.06 \\
\midrule
Fast-GraphRAG & 11.11 & 11.11 & 0.00 & 0.00 & 17.36 & 0.00 \\
HippoRAG2 & 44.44 & 44.44 & 0.00 & 10.00 & 3.61 & 2.50 \\
Microsoft GraphRAG (local) & 0.00 & 0.00 & 0.00 & 0.00 & 9.55 & 0.00 \\
Microsoft GraphRAG (global) & 5.56 & 5.56 & 0.00 & 0.00 & 7.61 & 0.00 \\
LightRAG (hybrid) & 27.78 & 27.78 & 0.00 & 8.33 & 11.56 & 2.22 \\
LinearRAG & 22.22 & 22.22 & 0.00 & 0.00 & 44.01 & 0.00 \\
\bottomrule
\end{tabular*}
\end{table}

\begin{table}[H]
\centering
\small
\label{tab:people}
\caption{People}

\begin{tabular*}{\linewidth}{@{\extracolsep{\fill}}ccccccc}
\toprule
\multirow[c]{2}{*}{\raisebox{-0.5ex}{\textbf{Method}}} &
\multicolumn{3}{c}{\textbf{Question Answering}} &
\multicolumn{3}{c}{\textbf{Summary}} \\
\cmidrule(lr){2-4}\cmidrule(lr){5-7}
& \textbf{Ave. Acc.} & \textbf{Single-fact Acc.} & \textbf{Multi-fact Acc.} & \textbf{Recall} & \textbf{Precision} & \textbf{F1} \\
\midrule
NaiveRAG & 62.38 & 76.62 & 28.12 & 10.48 & 15.29 & 8.03 \\
BM25 & 42.20 & 51.95 & 18.75 & 3.63 & 21.43 & 2.82 \\
\midrule
Fast-GraphRAG & 29.36 & 33.77 & 18.75 & 1.48 & 22.83 & 1.62 \\
HippoRAG2 & 62.39 & 72.73 & 37.50 & 7.63 & 15.69 & 6.14 \\
Microsoft GraphRAG (local) & 33.95 & 38.96 & 21.88 & 4.59 & 9.17 & 2.98 \\
Microsoft GraphRAG (global) & 55.05 & 62.34 & 37.50 & 5.52 & 14.13 & 5.41 \\
LightRAG (hybrid) & 58.71 & 67.53 & 37.50 & 5.56 & 15.69 & 4.73 \\
LinearRAG & 43.12 & 51.95 & 21.88 & 1.52 & 22.51 & 1.69 \\
\bottomrule
\end{tabular*}
\end{table}

\begin{table}[H]
\centering
\small
\label{tab:philosophy}
\caption{Philosophy}

\begin{tabular*}{\linewidth}{@{\extracolsep{\fill}}ccccccc}
\toprule
\multirow[c]{2}{*}{\raisebox{-0.5ex}{\textbf{Method}}} &
\multicolumn{3}{c}{\textbf{Question Answering}} &
\multicolumn{3}{c}{\textbf{Summary}} \\
\cmidrule(lr){2-4}\cmidrule(lr){5-7}
& \textbf{Ave. Acc.} & \textbf{Single-fact Acc.} & \textbf{Multi-fact Acc.} & \textbf{Recall} & \textbf{Precision} & \textbf{F1} \\
\midrule
NaiveRAG & 46.15 & 50.00 & 16.67 & 7.41 & 8.69 & 2.44 \\
BM25 & 26.92 & 28.26 & 16.67 & 1.39 & 6.17 & 0.79 \\
\midrule
Fast-GraphRAG & 50.00 & 47.83 & 66.67 & 2.96 & 12.71 & 1.01 \\
HippoRAG2 & 67.31 & 69.57 & 50.00 & 8.52 & 13.98 & 3.58 \\
Microsoft GraphRAG (local) & 46.16 & 43.48 & 66.67 & 5.74 & 4.44 & 1.73 \\
Microsoft GraphRAG (global) & 65.38 & 63.04 & 83.33 & 7.13 & 8.35 & 2.28 \\
LightRAG (hybrid) & 71.16 & 69.57 & 83.33 & 8.52 & 8.73 & 4.80 \\
LinearRAG & 50.00 & 50.00 & 50.00 & 1.85 & 17.79 & 2.56 \\
\bottomrule
\end{tabular*}
\end{table}

\begin{table}[H]
\centering
\small
\label{tab:religion}
\caption{Religion}

\begin{tabular*}{\linewidth}{@{\extracolsep{\fill}}ccccccc}
\toprule
\multirow[c]{2}{*}{\raisebox{-0.5ex}{\textbf{Method}}} &
\multicolumn{3}{c}{\textbf{Question Answering}} &
\multicolumn{3}{c}{\textbf{Summary}} \\
\cmidrule(lr){2-4}\cmidrule(lr){5-7}
& \textbf{Ave. Acc.} & \textbf{Single-fact Acc.} & \textbf{Multi-fact Acc.} & \textbf{Recall} & \textbf{Precision} & \textbf{F1} \\
\midrule
NaiveRAG & 61.84 & 59.72 & 100.00 & 12.33 & 15.56 & 6.83 \\
BM25 & 39.47 & 38.89 & 50.00 & 6.14 & 17.66 & 3.33 \\
\midrule
Fast-GraphRAG & 34.21 & 30.56 & 100.00 & 3.81 & 20.06 & 2.50 \\
HippoRAG2 & 67.11 & 68.06 & 50.00 & 8.87 & 19.88 & 5.94 \\
Microsoft GraphRAG (local) & 27.63 & 27.78 & 25.00 & 3.45 & 10.12 & 1.74 \\
Microsoft GraphRAG (global) & 50.00 & 47.22 & 100.00 & 8.73 & 13.37 & 2.65 \\
LightRAG (hybrid) & 60.52 & 58.33 & 100.00 & 9.48 & 19.95 & 5.47 \\
LinearRAG & 46.05 & 44.44 & 75.00 & 0.48 & 29.37 & 0.69 \\
\bottomrule
\end{tabular*}
\end{table}

\begin{table}[H]
\centering
\small
\label{tab:society}
\caption{Society}

\begin{tabular*}{\linewidth}{@{\extracolsep{\fill}}ccccccc}
\toprule
\multirow[c]{2}{*}{\raisebox{-0.5ex}{\textbf{Method}}} &
\multicolumn{3}{c}{\textbf{Question Answering}} &
\multicolumn{3}{c}{\textbf{Summary}} \\
\cmidrule(lr){2-4}\cmidrule(lr){5-7}
& \textbf{Ave. Acc.} & \textbf{Single-fact Acc.} & \textbf{Multi-fact Acc.} & \textbf{Recall} & \textbf{Precision} & \textbf{F1} \\
\midrule
NaiveRAG & 62.07 & 74.24 & 23.81 & 17.31 & 19.22 & 12.89 \\
BM25 & 28.73 & 33.33 & 14.29 & 8.64 & 18.00 & 5.58 \\
\midrule
Fast-GraphRAG & 54.02 & 59.09 & 38.10 & 12.41 & 24.93 & 9.55 \\
HippoRAG2 & 68.97 & 78.79 & 38.10 & 13.64 & 13.45 & 8.92 \\
Microsoft GraphRAG (local) & 50.57 & 56.06 & 33.33 & 17.38 & 10.12 & 7.70 \\
Microsoft GraphRAG (global) & 65.52 & 71.21 & 47.62 & 14.78 & 12.39 & 8.02 \\
LightRAG (hybrid) & 66.67 & 75.76 & 38.10 & 13.70 & 14.49 & 9.91 \\
LinearRAG & 50.57 & 56.06 & 33.33 & 8.77 & 28.99 & 10.14 \\
\bottomrule
\end{tabular*}
\end{table}

\begin{table}[H]
\centering
\small
\label{tab:technology}
\caption{Technology}

\begin{tabular*}{\linewidth}{@{\extracolsep{\fill}}ccccccc}
\toprule
\multirow[c]{2}{*}{\raisebox{-0.5ex}{\textbf{Method}}} &
\multicolumn{3}{c}{\textbf{Question Answering}} &
\multicolumn{3}{c}{\textbf{Summary}} \\
\cmidrule(lr){2-4}\cmidrule(lr){5-7}
& \textbf{Ave. Acc.} & \textbf{Single-fact Acc.} & \textbf{Multi-fact Acc.} & \textbf{Recall} & \textbf{Precision} & \textbf{F1} \\
\midrule
NaiveRAG & 57.30 & 64.29 & 45.45 & 10.87 & 24.03 & 11.91 \\
BM25 & 40.45 & 44.64 & 33.33 & 6.11 & 22.38 & 5.45 \\
\midrule
Fast-GraphRAG & 17.98 & 17.86 & 18.18 & 2.22 & 33.77 & 2.94 \\
HippoRAG2 & 59.55 & 66.07 & 48.48 & 6.91 & 18.85 & 4.92 \\
Microsoft GraphRAG (local) & 43.82 & 39.29 & 51.52 & 7.57 & 23.76 & 7.56 \\
Microsoft GraphRAG (global) & 47.19 & 44.64 & 51.52 & 8.40 & 19.96 & 7.10 \\
LightRAG (hybrid) & 43.82 & 46.43 & 39.39 & 9.55 & 21.26 & 8.68 \\
LinearRAG & 47.19 & 48.21 & 45.45 & 2.22 & 36.39 & 3.00 \\
\bottomrule
\end{tabular*}
\end{table}

\section{Prompts for Data Construction}
\label{sec:appendix_prompts}

\tcbset{
    promptbox/.style={
        colback=gray!8,
        colframe=gray!50,
        coltitle=white,
        fonttitle=\bfseries\sffamily,
        colbacktitle=gray!70!black,
        boxrule=0.5pt,
        arc=3pt,
        left=8pt,
        right=8pt,
        top=6pt,
        bottom=6pt,
        width=0.9\linewidth,
        before upper={\small}
    }
}

\begin{figure}[H]
\centering
\begin{tcolorbox}[promptbox, title=Question Generation: Single-fact]
You are constructing a question for a \textbf{SINGLE-FACT} (supported by one citation) citation-based QA dataset.

\textbf{ARTICLE TITLE:} \{\{WIKI\_TITLE\}\}

\textbf{SECTION PATH:} \{\{SECTION\_PATH\}\}

\textbf{WIKI SENTENCE} (with inline citations): \{\{SENTENCE\}\}

\textbf{CLEAN FACTUAL STATEMENT} (this will be used as the reference answer): \{\{STATEMENT\}\}

\textbf{REFERENCE URLS} (for context, do NOT quote them explicitly): \{\{REF\_URLS\_LIST\}\}

\textbf{Your task:}
\begin{itemize}[leftmargin=*, nosep]
\item Write ONE natural-language QUESTION in English or Chinese (depending on the style of the article), such that:
  \begin{itemize}[leftmargin=*, nosep]
  \item The gold answer should be exactly the given STATEMENT (possibly with tiny paraphrasing).
  \item The question should contain \textbf{multiple constraints} (e.g. entity + time, quantity + condition, entity + location).
  \item If any of these constraints was removed, the question would become under-specified or wrong.
  \item The question must be answerable solely from the given statement and sentence.
  \end{itemize}
\item The question should feel natural and non-trivial:
  \begin{itemize}[leftmargin=*, nosep]
  \item Do NOT copy any span of 4 or more consecutive words from the sentence or the statement.
  \item Avoid generic patterns like ``What is X?'', ``Who is Y?'', ``When did X happen?''.
  \end{itemize}
\end{itemize}

\textbf{Return JSON ONLY:} \texttt{\{"question": "..."\}}
\end{tcolorbox}
\caption{Prompt for Single-fact question generation}
\label{fig:prompt1}
\end{figure}

\begin{figure}[H]
\centering
\begin{tcolorbox}[promptbox, title=Question Generation: Multi-fact]
You are constructing a question for a \textbf{MULTI-FACT} (requires several citations together) citation-based QA dataset.

\textbf{ARTICLE TITLE:} \{\{WIKI\_TITLE\}\}

\textbf{SECTION PATH:} \{\{SECTION\_PATH\}\}

\textbf{WIKI SENTENCE} (with inline citations): \{\{SENTENCE\}\}

\textbf{CLEAN FACTUAL STATEMENT} (this will be used as the reference answer): \{\{STATEMENT\}\}

\textbf{REFERENCE URLS} (for context, do NOT quote them explicitly): \{\{REF\_URLS\_LIST\}\}

\textbf{Your task:}
\begin{itemize}[leftmargin=*, nosep]
\item Write ONE natural-language QUESTION in English or Chinese (depending on the style of the article), such that:
  \begin{itemize}[leftmargin=*, nosep]
  \item The gold answer should be exactly the given STATEMENT (possibly with tiny paraphrasing).
  \item The question should contain \textbf{multiple constraints} (e.g. entity + time, quantity + condition, entity + location).
  \item If any of these constraints was removed, the question would become under-specified or wrong.
  \item The question must be answerable solely from the given statement and sentence.
  \end{itemize}
\item The question should feel natural and non-trivial:
  \begin{itemize}[leftmargin=*, nosep]
  \item Do NOT copy any span of 4 or more consecutive words from the sentence or the statement.
  \item Avoid generic patterns like ``What is X?'', ``Who is Y?'', ``When did X happen?''.
  \end{itemize}
\end{itemize}

\textbf{Return JSON ONLY:} \texttt{\{"question": "..."\}}
\end{tcolorbox}
\caption{Prompt for Multi-fact question generation.}
\label{fig:prompt2}
\end{figure}

\begin{figure}[H]
\centering
\begin{tcolorbox}[promptbox, title=Question Generation: Summary]
You are constructing a \textbf{TOPIC-CENTERED SUMMARY QUESTION} for a topic.

\textbf{TOPIC PATH} (broad -> specific): \{\{SECTION\_PATH\}\}

\textbf{OPTIONAL BODY EXCERPT} (for natural phrasing only): \{\{BODY\_EXCERPT\}\}

\textbf{GOLD STATEMENTS} (facts that a good answer SHOULD cover; do NOT quote them): \{\{GOLD\_STATEMENTS\_LIST\}\}

\textbf{Your task:}
\begin{itemize}[leftmargin=*, nosep]
\item Write ONE natural-language question that asks for a concise, encyclopedic-style overview of the MOST SPECIFIC topic (typically the LAST 1--2 elements of the path).
\item Use the GOLD STATEMENTS only as soft guidance to choose what aspects to emphasize, so that the answer naturally tends to cover those facts.
\item The question must remain strongly anchored to the leaf topic in the path.
\end{itemize}

\textbf{STRICT constraints:}
\begin{itemize}[leftmargin=*, nosep]
\item DO NOT mention Wikipedia/article/section/heading or similar meta words.
\item DO NOT copy any span of 4+ consecutive words from any gold statement.
\item Avoid leaking specific factual details from the gold statements in the question (especially exact numbers, exact dates, long proper names, or verbatim event descriptions).
\item You may mention high-level aspects (e.g., ``history'', ``structure'', ``major components'', ``development'', ``reception'') if they align with the leaf topic and the gold statements.
\item 20--200 characters.
\end{itemize}

\textbf{Return JSON ONLY:} \texttt{\{"question": "..."\}}
\end{tcolorbox}
\caption{Prompt for Summary question generation.}
\label{fig:prompt3}
\end{figure}

\begin{figure}[H]
\centering
\begin{tcolorbox}[promptbox, title=Summary Question Filtering]
You are doing a \textbf{POST-HOC VERIFICATION} for a citation-based summary dataset.

\textbf{ARTICLE TITLE:} \{\{WIKI\_TITLE\}\}

\textbf{LEAF SECTION TOPIC PATH:} \{\{SECTION\_PATH\}\}

\textbf{Leaf topic} (most specific): \{\{LEAF\_TOPIC\}\}

You will be given multiple ITEMS. Each item has:
\begin{itemize}[leftmargin=*, nosep]
\item a STATEMENT (candidate gold statement)
\item several REFERENCES (content excerpts)
\end{itemize}

\textbf{Task:} For EACH item:
\begin{itemize}[leftmargin=*, nosep]
\item keep=true only if the REFERENCES (collectively) contain enough information to support ALL key factual claims in the STATEMENT.
\item keep=false if key facts are missing, contradicted, or the references are irrelevant/noisy.
\end{itemize}

\textbf{Rules:}
\begin{itemize}[leftmargin=*, nosep]
\item Use ONLY the given references; ignore outside knowledge.
\item Be fairly strict: if unsure due to missing evidence, set keep=false.
\end{itemize}

\textit{\# [Items are dynamically inserted here]}

\textbf{Return JSON ONLY:} \texttt{\{"items":[{"idx":1,"keep":true/false,"reason":"brief"}], "summary":"brief"\}}
\end{tcolorbox}
\caption{Prompt for summary question filtering.}
\label{fig:prompt4}
\end{figure}

\begin{figure}[H]
\centering
\begin{tcolorbox}[promptbox, title=Single-fact Question Filtering]
You are checking whether the provided REFERENCES collectively support a Q\&A.

\textbf{Q:} \{\{QUESTION\}\}

\textbf{A:} \{\{ANSWER\}\}

\textbf{REFERENCES} (may include some noise, read holistically): \{\{REFERENCES\_CONTENT\_BUNDLE\}\}

\textbf{Rules:}
\begin{itemize}[leftmargin=*, nosep]
\item If the references together contain the key facts to justify the answer, return supported=true.
\item If key facts are missing or contradicted, return supported=false.
\end{itemize}

\textbf{Return JSON ONLY} (do NOT explain your reasoning process, be concise):

\texttt{\{"supported": true/false, "reason": "brief"\}}
\end{tcolorbox}
\caption{Prompt for Single-fact question filtering.}
\label{fig:prompt5}
\end{figure}

\begin{figure}[H]
\centering
\begin{tcolorbox}[promptbox, title=Multi-fact Question Filtering]
You are given a factual STATEMENT (used as the reference answer for a QA pair), together with the QUESTION and several reference documents cited from Wikipedia.

\textbf{QUESTION:} \{\{QUESTION\}\}

\textbf{REFERENCE ANSWER (STATEMENT):} \{\{STATEMENT\}\}

\textbf{REFERENCES:} \{\{REFERENCES\_CONTENT\_BUNDLE\}\}

Your task is to judge whether these references are \textbf{jointly necessary} to support the FULL factual content of the STATEMENT.

\textbf{Rules:}
\begin{enumerate}[leftmargin=*, nosep]
\item Consider ONLY the information contained in the given references. Ignore any outside world knowledge.
\item For EACH reference individually, imagine you only had that single reference:
  \begin{itemize}[leftmargin=*, nosep]
  \item If that single reference ALONE already contains enough information to support ALL key factual claims in the STATEMENT (numbers, named entities, relationships, important conditions), then that reference is ``individually sufficient'' to justify the STATEMENT.
  \end{itemize}
\item If \textbf{ANY} single reference is individually sufficient, then the multi-reference pattern is NOT truly necessary. $\rightarrow$ In this case, set all\_needed = false.
\item Only if \textbf{NO} single reference is individually sufficient (each one misses some essential facts), and you really need to COMBINE at least two references to cover the full STATEMENT, set all\_needed = true.
\end{enumerate}

``Key factual claims'' means the main facts expressed by the STATEMENT, not minor stylistic details.

\textbf{Return JSON ONLY:} \texttt{\{"all\_needed": true/false, "reason": "brief"\}}
\end{tcolorbox}
\caption{Prompt for Multi-fact question filtering.}
\label{fig:prompt6}
\end{figure}

\end{document}